\def\paperID{}
\def\confName{}
\def\confYear{2025}

\def\authorBlock{
	Zixiao Gu$^1$, ~
	Mengtian Li\footnotemark[1] $^,$$^1$$^,$$^2$, ~
	Ruhua Chen$^3$, ~
	Zhongxia Ji$^3$, \\
	Sichen Guo$^3$, ~
	Zhenye Zhang$^1$, ~
	Guangnan Ye\footnotemark[1] $^,$$^1$, ~
	Zuo Hu$^3$ \\
	$^1$Fudan University, \\
	$^2$Shanghai University, \\
	$^3$Shanghai Theatre Academy \\
	{\tt\small zxgu23@m.fudan.edu.cn}
	{\tt\small yegn@fudan.edu.cn}
}

\newif\ifreview 
\newif\ifarxiv \newcommand{\arxiv}{\arxivtrue}
\newif\ifcamera 
\newif\ifrebuttal 
\arxiv 

\pdfoutput=1
\documentclass[10pt,twocolumn,letterpaper]{article}
\ifreview \usepackage[review]{cvpr} \fi
\ifarxiv \usepackage[pagenumbers]{cvpr} \fi
\ifrebuttal \usepackage[rebuttal]{cvpr} \fi
\ifcamera \usepackage{cvpr} \fi

\usepackage{graphicx}	
\usepackage{amsmath}	
\usepackage{amssymb}	
\usepackage{booktabs}
\usepackage{times}
\usepackage{microtype}
\usepackage{epsfig}
\usepackage{caption}
\usepackage{float}
\usepackage{placeins}
\usepackage{color, colortbl}
\usepackage{stfloats}
\usepackage{enumitem}
\usepackage{tabularx}
\usepackage{xstring}
\usepackage{multirow}
\usepackage{xspace}
\usepackage{url}
\usepackage{subcaption}
\usepackage{xcolor}
\usepackage[hang,flushmargin]{footmisc}

\ifcamera \usepackage[accsupp]{axessibility} \fi

% ...

\ifarxiv  \fi

\newcommand{\R}[1]{{%
		\textbf{%
			\ifstrequal{#1}{1}{\textcolor{red}{R#1}}{%
				\ifstrequal{#1}{2}{\textcolor{blue}{R#1}}{%
					\ifstrequal{#1}{3}{\textcolor{magenta}{R#1}}{%
						\ifstrequal{#1}{4}{\textcolor{teal}{R#1}}{%
							\textcolor{cyan}{R#1}%
			}}}}%
		}%
}}  

\usepackage{xr-hyper}

\makeatletter
\newcommand*{\addFileDependency}[1]{
	\typeout{(#1)}
	\@addtofilelist{#1}
	\IfFileExists{#1}{}{\typeout{No file #1.}}
}

\makeatother
\newcommand*{\myexternaldocument}[1]{
	\externaldocument{#1}
	\addFileDependency{#1.tex}
	\addFileDependency{#1.aux}
}

\definecolor{cvprblue}{rgb}{0.21,0.49,0.74}
\usepackage[pagebackref,breaklinks,colorlinks,allcolors=cvprblue]{hyperref}
\usepackage[capitalize]{cleveref}
\crefname{section}{Sec.}{Secs.}
\crefname{table}{Table}{Tables}
\crefname{figure}{Fig.}{Figs.}

\ifarxiv \crefname{appendix}{App.}{Apps.}
\else \crefname{appendix}{Suppl.}{Suppls.} \fi

\unless\ifarxiv \myexternaldocument{_supplementary} \fi

\begin{document}
	\title{ArtNVG: Content-Style Separated Artistic Neighboring-View Gaussian Stylization}
	\author{\authorBlock}
	
	\twocolumn[{
		\renewcommand\twocolumn[1][]{#1}
		\maketitle
		\begin{center}
			\captionsetup{type=figure}
			\includegraphics[width=\textwidth]{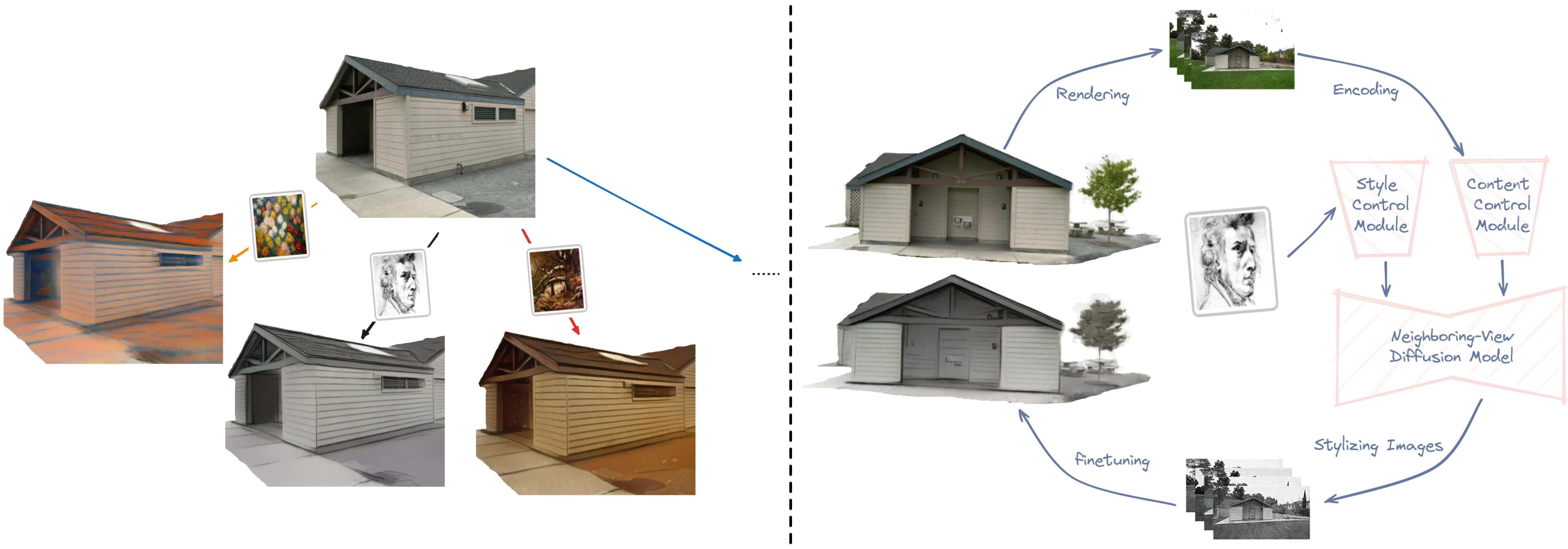}
			\captionof{figure}{\textbf{ArtNVG.}
				Our method apply the style of a reference image to a 3D Gaussian Splatting (3DGS) scene (Left).
				This is realized by stylizing the renderings of 3DGS and finetuning the original scene (Right).
				Our contribution is a content-style separated neighboring-view aligned stylization framework, which primarily reduce the risk of information leakage and improve consistency of local colors and textures.
			}
			\label{fig:teaser}
		\end{center}
	}]

	\maketitle
	\begin{abstract}
As demand from the film and gaming industries for 3D scenes with target styles grows, the importance of advanced 3D stylization techniques increases.
However, recent methods often struggle to maintain local consistency in color and texture throughout stylized scenes, which is essential for maintaining aesthetic coherence.
To solve this problem, this paper introduces ArtNVG, an innovative 3D stylization framework that efficiently generates stylized 3D scenes by leveraging reference style images.
Built on 3D Gaussian Splatting (3DGS), ArtNVG achieves rapid optimization and rendering while upholding high reconstruction quality.
Our framework realizes high-quality 3D stylization by incorporating two pivotal techniques: Content-Style Separated Control and Attention-based Neighboring-View Alignment.
Content-Style Separated Control uses the CSGO model and the Tile ControlNet to decouple the content and style control, reducing risks of information leakage.
Concurrently, Attention-based Neighboring-View Alignment ensures consistency of local colors and textures across neighboring views, significantly improving visual quality.
Extensive experiments validate that ArtNVG surpasses existing methods, delivering superior results in content preservation, style alignment, and local consistency.
\end{abstract}

	\section{Introduction}
\label{sec:intro}


Propelled by the advancement of 3D reconstruction techniques, capturing and representing 3D scenes of rich details and impressive photorealism has become increasingly accessible and efficient.
However, there still exist challenges in creating 3D stylized scenes, which is an immense demand in the film and game industry.
One way to address this problem is 3D scene stylization and the goal of 3D scene stylization is to apply the specific style from a given reference image to a constructed 3D scene.

With the remarkable development of radiance fields \cite{mildenhall2021nerf} in 3D reconstruction, it is more convenient for us to capture 3D scenes.
Given a collection of images and their corresponding camera parameters (typically from a structure-from-motion system, such as COLMAP \cite{schoenberger2016sfm}), one can easily construct a 3D scene by using radiance fields.
Compared to other 3D representations, the implicit and differentiable nature of NeRF makes it more suitable to do stylization tasks under the style guidance of high-level semantic information.
Thus, it has caught significant attention from researchers worldwide and many radiance fields-based 3D stylization methods have emerged \cite{zhang2022arf,zhang2024coarf}.
However, the slow training and rendering speed of radiance fields restricts its application, although there have been several works \cite{muller2022instant,fridovich2022plenoxels} concentrating on relieving this problem.

Recently, a newly introduced 3D representation, 3D Gaussian Splatting (3DGS) \cite{kerbl20233d} enables a fast training and rendering speed while promising the high quality of reconstructed scenes.
Also, 3DGS can be seamlessly integrated into current deep-learning-based 3D stylization frameworks, allowing for real-time 3D stylization \cite{jain2024stylesplat,liu2024stylegaussian,yu2024instantstylegaussian}.
Benefited by the excellent performance diffusion models have achieved in 2D editing and stylization, researchers have explored using diffusion-based 2D stylization methods to facilitate 3D editing and stylization tasks \cite{igs2gs,wu2024gaussctrl,fujiwara2024style,yu2024instantstylegaussian}.
Despite the better effects they achieve than previous works, they still suffer from the problem of information leakage.
They edit or stylize an image in a way like SDEdit \cite{mengsdedit}, which is likely to leak the style information of the content image and mixes the content and style controls.
Meanwhile, the performance of these methods is degraded due to local inconsistency among generated stylized images, because images of neighboring views are processed independently without taking the largely overlapped regions into account.

To solve the above problems, we propose a 3D stylization pipeline ArtNVG, allowing for rapid stylization of existing 3DGS scenes with the target style images. This framework ensures the style alignment of the given image while maintaining the original scene's content semantics with locally consistent colors and textures.
Our method captures renderings from the original 3DGS scene as content images.
Then, we use the projection modules from CSGO \cite{xing2024csgo} to encode the content images and the style image into controls separately.
Besides, we group neighboring views to fully share the local information among neighboring views.
The content and style controls are passed to cross-attention layers, while information among neighboring views is shared in self-attention layers.
Finally, we use the stylized images to finetune the original scene.

Extensive experiments validate that our method achieves superior 3D stylization with high-quality results, offering improvements in content preservation, style alignment, and local consistency compared to previous 3D stylization methods.
The contributions of our work can be summarized in three aspects:

\begin{enumerate}
	\item We propose a zero-shot 3D stylization method, allowing for applying the style of a given image to an existing 3DGS scene. This pipeline enables fast optimization and rendering with no need for extra pre-training.
	\item We employ Content-Style Separated Control to reduce the risk of information leakage.
	\item We design an Attention-based Neighboring-View Alignment mechanism to maintain local consistency.
\end{enumerate}
	\section{Related Work}
\label{sec:related}
\subsection{3D Representations}
\label{3d-representations}
In the field of 3D scene reconstruction and editing, various representations have been proposed to address corresponding tasks.
Among these, NeRF \cite{mildenhall2021nerf} have made significant strides in this area, employing volumetric rendering to do 3D scene reconstruction.
Researchers have explored a wide range of applications of NeRF across different domains of computer vision and graphics, such as novel view synthesis \cite{barron2021mip,barron2022mip,pumarola2021d}, 3D generation \cite{chan2022efficient,gu2021stylenerf,niemeyer2021giraffe,schwarz2020graf} and surface reconstruction \cite{wang2021neus,wang2023neus2}.
There are several variants of radiance fields with different implementation methods, including MLPs \cite{barron2021mip,barron2022mip,mildenhall2021nerf}, decomposed tensors \cite{chan2022efficient,chen2022tensorf,fridovich2023k}, hash tables \cite{muller2022instant} and voxels \cite{fridovich2022plenoxels,sun2022direct}.
However, despite its high rendering quality, NeRF is very time-consuming in reconstruction \cite{barron2021mip,barron2022mip,muller2022instant} and editing \cite{haque2023instruct,zhang2022arf}, because of its reliance on dense volumetric sampling.

Recently, 3DGS \cite{kerbl20233d} has emerged as a more efficient alternative for its capabilities of rapid construction, real-time rendering, and exceptional rendering quality.
Its excellent characteristics make it popular in diverse domains.
For example, it has been employed in generation \cite{chen2024text,yi2023gaussiandreamer} and editing \cite{wu2024gaussctrl,wang2024gaussianeditor}.
In this work, we apply 3DGS as our 3D representation for fast stylization and rendering.

\subsection{Diffusion-based 3D Editing}
\label{db3dediting}

Recent advancements in 3D scene editing have increasingly leveraged diffusion-based methods to enhance fidelity and control.
Instruct-NeRF2NeRF (IN2N) \cite{haque2023instruct} utilizes InstructPix2Pix to iteratively refine the edits on a 3D scene as per text instructions.
This method exploits the potential of diffusion models for detailed guidance, ensuring that the model’s outputs progressively align with the user's textual descriptions.
However, since consistent editing of multi-view images is not guaranteed, it faces challenges such as slow processing speed and inconsistency across different views.
ViCA-NeRF \cite{dong2024vica} follows a similar diffusion-based approach to IN2N, selecting reference images from the dataset to improve consistency and address the problem of blurry editing.
Instruct-GS2GS (IG2G) \cite{igs2gs} is an updated version of IN2N by using 3DGS.
GaussianEditor \cite{wang2024gaussianeditor,yi2023gaussiandreamer} further expands on this paradigm by integrating segmentation, enhancing the selectivity and precision of edits.
Additionally, GaussCtrl \cite{wu2024gaussctrl} utilizes SAM \cite{kirillov2023segment} to extract more precise masks for accurate editing. In addition, Gaussctrl proposes a novel framework that integrates the depth-conditioned ControlNet with the attention-based latent code alignment to improve multi-view consistency.

\subsection{3D Stylization}
\label{3D stylization}
As the demands for stylized 3D contents continue to grow, the field of neural stylization has significantly broadened to encompass a variety of 3D forms.
Early researches \cite{kato2018neural,michel2022text2mesh,yin20213dstylenet} often involved differential rendering to transfer styles onto 3D meshes, facilitating both geometric and texture adaptations that align 3D objects with specific artistic styles.
Subsequent advancements employed point clouds as the 3D representation to ensure continuity and consistency across different viewpoints during stylization.
For example, LSNV \cite{huang2021learning} uses featurized 3D point clouds combined with a 2D CNN renderer, allowing for the generation of stylized renderings.
However, explicit methods are often limited by the quality of the 3D reconstruction, which can lead to noticeable artifacts in detailed and complex real world scenes.

In response to these limitations, implicit methods based on NeRF have emerged to enhance both the accuracy and aesthetic quality of stylization for their excellent quality of reconstruction.
Many NeRF-based methods \cite{huang2022stylizednerf,nguyen2022snerf,zhang2022arf} have significantly pushed the boundaries of 3D stylization by incorporating losses specifically tailored for image style transfer within their training processes, enabling more precise control over the color fidelity relative to the style image.
Some approaches \cite{wang2023nerf,xu2023desrf} involve geometric stylization to mimic the reference style, promising consistency in novel view stylization.
Even though these methods achieve good stylization, they still suffer from time-consuming optimization and slow rendering speed, inherited from NeRF.

To accelerate the training and rendering speed of 3D stylization, 3DGS becomes an alternative for researchers \cite{liu2024stylegaussian,saroha2024gaussian,zhang2024stylizedgs,jain2024stylesplat}.
StyleGaussian \cite{liu2024stylegaussian}, following the idea of StyleRF \cite{liu2023stylerf}, proposes a feedforward-based framework, which uses AdaIN \cite{karras2019style} to capture style features.
StyleSplat \cite{jain2024stylesplat} is a segmentation-based method, using NNFM loss to do stylization.
StylizedGS \cite{zhang2024stylizedgs} is a filter-based method, accompanied by several forms of user control.
Restricted by ability of style transfer methods they use, their stylized renderings often show locally inconsistent colors and textures.

Motivated by diffusion-based 3D editing methods \cite{haque2023instruct,dong2024vica,igs2gs,wu2024gaussctrl}, recent methods \cite{fujiwara2024style,yu2024instantstylegaussian} have explored the use of diffusion models in 3D stylization.
Improvement in stylization though they achieve, yet they still have problems with information leakage and local inconsistency.
Our work is also based on diffusion models, but reduces the risk of information leakage and promises local consistency.
	\section{Method}
\label{sec:method}
\begin{figure*}[htbp]
	\centering
	\includegraphics[width=\linewidth]{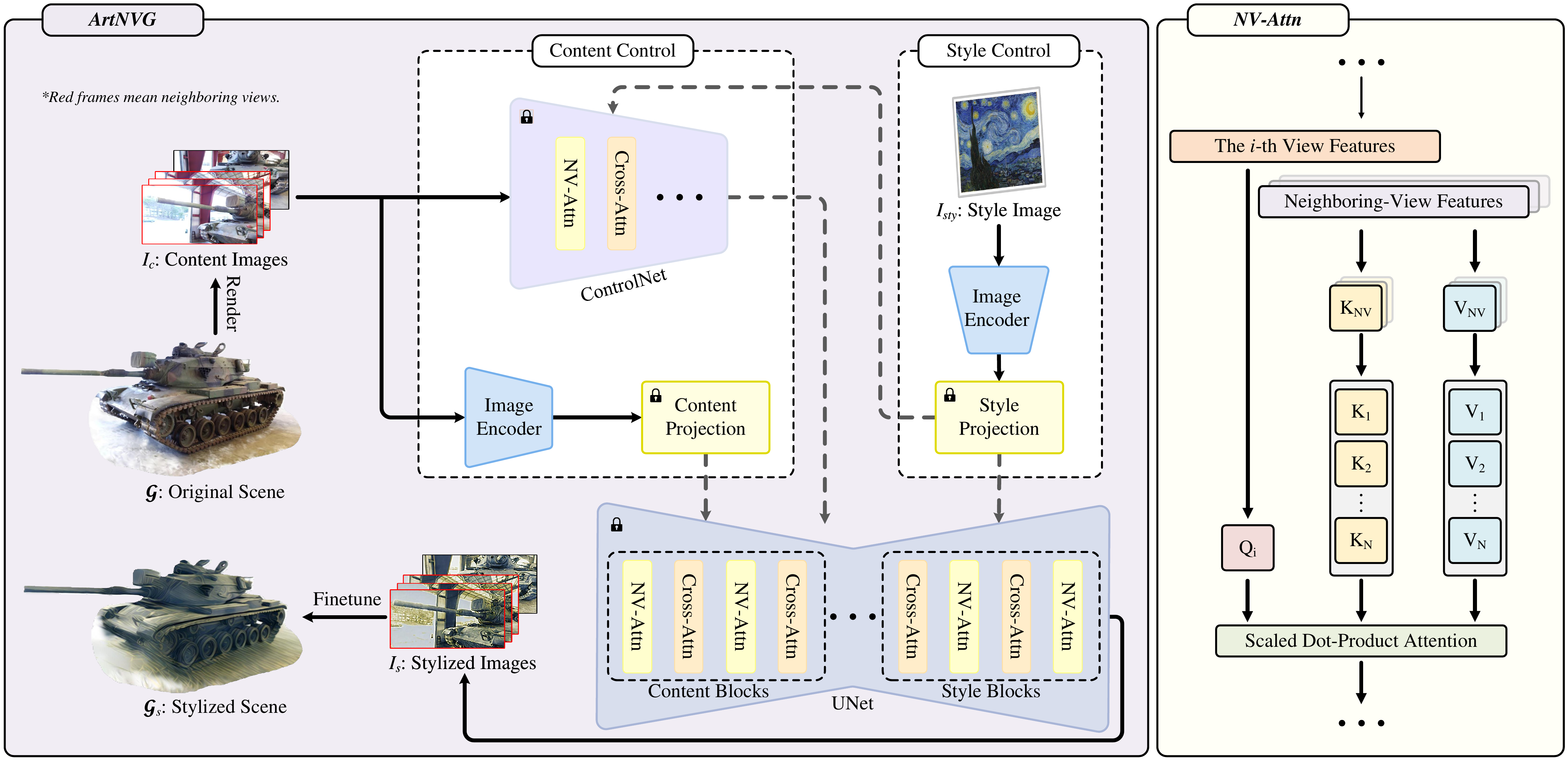}
	\caption{(a) Left: Overview of ArtNVG. (b) Right: Architecture of the Neighboring-View Attention Layer. Given an original 3DGS scene $\mathcal G$, we first render the content images $I_c$ from it. Then, we encode the content images $I_c$ and style image $I_\mathrm{sty}$ separately to get the content and style control. Meanwhile, neighboring views of content images are clustered into groups and sampled together in the Neighboring-View Diffusion Model. Finally, stylized images are used to finetune $\mathcal G$ and get the stylized 3DGS scene $\mathcal G_s$.}
	\label{fig:pipeline}
\end{figure*}
\subsection{Preliminary}
\label{sec:preliminary}
\textbf{3D Gaussian Splatting.}
3DGS \cite{kerbl20233d} is a 3D representation technique, by which the scene is modeled by a collection of 3D Gaussians. 
Each Gaussian is characterized by its spatial location $ \mu $ and its full covariance matrix $ \Sigma $: 
\begin{equation}
	\mathcal G(x) = \mathrm e^{-\frac{1}{2} (x - \mu)^\mathsf{T} \Sigma^{-1} (x - \mu)}.
\end{equation}
During the rendering process, 3D Gaussians are projected onto 2D planes using a transformation that adjusts their covariance matrices to the viewing plane.
This projection involves converting the covariance matrix $ \Sigma $ to a new covariance matrix $ \Sigma' $:
\begin{equation}
	\Sigma' = J W \Sigma W^\mathsf{T} J^\mathsf{T},
\end{equation} 
where $ W $ denotes the viewing transformation matrix and $ J $ is the Jacobian of the affine approximation of the projective transformation.
To support differentiable optimization, the covariance matrix $ \Sigma $ is decomposed into a scaling matrix $ S $ and a rotation matrix $ R $, allowing for gradient-based learning of these parameters:
\begin{equation}
	\Sigma = R S S^\mathsf{T} R^\mathsf{T}.
\end{equation}
Our method realizes quick stylization of 3D scenes based on 3DGS representation.

\noindent\textbf{CSGO.}
CSGO \cite{xing2024csgo} introduces a novel approach to style transfer by effectively decoupling content and style features. This method optimizes the process of content-style composition built on top of Stable Diffusion (SD) \cite{rombach2022high}, enabling precise control over the integration of style elements while preserving the integrity of the original content. 
We employ CSGO for its ability to separate content and style features.

\noindent\textbf{NNFM Loss.}
NNFM Loss, proposed by Artistic Radiance Fields (ARF) \cite{zhang2022arf}, is designed to fuse 3D scenes with 2D style images by leveraging a VGG encoder.
This loss function enhances the transfer of high-frequency visual details from 2D style images to 3D scenes, ensuring the preservation of intricate texture details.
It computes the cosine distance between feature vectors from corresponding pixels in the style and rendered images.
For a feature vector at position $(i,j)$ in the feature map, $F_r$ and $F_s$ are feature vectors of the scene renderings and the style image, the loss function can be written as
\begin{equation}
	\mathcal L_\mathrm{NNFM} = \frac{1}{M} \sum_{i,j} \min \mathrm{D_{cos}}(F_r(i,j), F_s(i,j)).
\end{equation}
Here, $M$ is the number of pixels in the rendered image $F_r$ and $\mathrm{D_{cos}}$ denotes the cosine distance between two vectors. This method not only retains the 3D scene’s structural integrity but also infuses it with the artistic feature of the style image.

\subsection{ArtNVG}
\label{sec:css-anvg}
In this section, we describe our proposed 3D stylization framework: ArtNVG.
We will give an overview of our framework first (also illustrated in \cref{fig:pipeline}), followed by detailed explanations of every module.

\subsubsection{Overview}
\label{sec:method-overview}
\
\newline
Given a reference image and a 3D scene constructed from a collection of captured images and their corresponding camera poses, our task is to apply the specific style from the reference image to the 3D scene. 

Our method starts with rendering content images from the original 3DGS scene, followed by the Content-Style Separated Control, which encodes the content images and style image separately using the projection modules from CSGO \cite{xing2024csgo} and a Tile ControlNet \cite{zhang2023adding}.
Simultaneously, we sample and group neighboring views to ensure local consistency.
During the denoising process, the down blocks of the UNet mainly control the content, whereas the up blocks control the style.
Thus, the content and style controls are passed to different cross-attention layers. 
As for information among neighboring views, it is shared in self-attention layers in both the UNet and the Tile ControlNet, so we replace the original self-attention layers with the Neighboring-View Attention layers to maintain neighboring view consistency.
Finally, stylized images output by the neighboring-view diffusion model are used to finetune the original 3D scene and we apply NNFM loss proposed by Artistic Radiance Fields (ARF) \cite{zhang2022arf} to enhance style alignment and details in stylization.
\subsubsection{Content-Style Separated Control}
\label{sec:css-control}
\
\newline
To stylize a 3D scene, we need to ensure effective style transfer and accurate content preservation.
In 3D stylization tasks, one of the main concerns for everyone is the risk of content information leakage of the style image and style information leakage of content images.
Previous methods \cite{fujiwara2024style,yu2024instantstylegaussian} use an inversion-based content preservation approach, which adds noises to the latents of original renderings as the input of the diffusion model, posing a high risk of content information leakage.
To reduce the risk of unwanted information leakage, we encode the style image and content images separately and inject controls through cross-attention layers.

\noindent\textbf{Style Control Module.}
Our method uses the CSGO style projection module as the style control module.
The style control can be denoted as $\mathcal D_s=\mathcal F_s(\mathcal E(I_\mathrm{sty}))$, where $I_\mathrm{sty}$ is the style image, $\mathcal E$ is the CLIP image encoder, and $\mathcal{F}_s$ is the CSGO style projection module.

\noindent\textbf{Content Control Module.} 
We use the CSGO content projection module and a Tile ControlNet as the content control module, which accepts the content image $I_c$ rendered from the original scene $\mathcal G$ as the input.
The content control can be denoted as $\mathcal F_c(z_c^t, I_c, \mathcal D_s)=\mathcal F_c^{\prime}(z_c^t, \mathcal E(I_c))+\mathcal C(z_c^t, \mathcal E(I_c), \mathcal D_s)$, where $z_c^t$ is the latent code of the content image $I_c$ at time step $t$, $\mathcal{F}^{\prime}_c$ is the CSGO content projection module, and $\mathcal C$ is the Tile ControlNet.
The style control $\mathcal D_s$ needs to be injected into the Tile ControlNet as condition, because the ControlNet has a risk of leaking style information of content images, resulting in damaged results with low style alignment.

The outputs of two modules $\mathcal F_c(z_c^t, I_c, \mathcal D_s)$, $\mathcal D_s$ are passed to the UNet module $\mathcal{F}_{\mathrm{UNet}}$ from cross attention layers.
Then, we can denote the diffusion process as:
\begin{align}
	\epsilon^t &= \mathcal{F}_{\mathrm{UNet}}(z_c^t,t,\mathcal F_c(z_c^t, I_c, \mathcal D_s),\mathcal D_s),\label{eq:prediction}\\
	z_c^{t+1} &= \sqrt{\alpha_{t+1}}\frac{z_c^t - \sqrt{1-\alpha_t} \cdot \epsilon^t}{\sqrt{\alpha_t}} + \sqrt{1-\alpha_{t+1}} \epsilon^t,\label{eq:denoising}
\end{align}
where $\epsilon^t$ is the predicted noise and $\alpha_t$ is the scheduling coefficient in DDIM scheduler.

\subsubsection{Attention-based Neighboring-View Alignment}
\label{sec:method-NV}
\begin{figure*}[htbp]
	\centering
	\includegraphics[width=\linewidth]{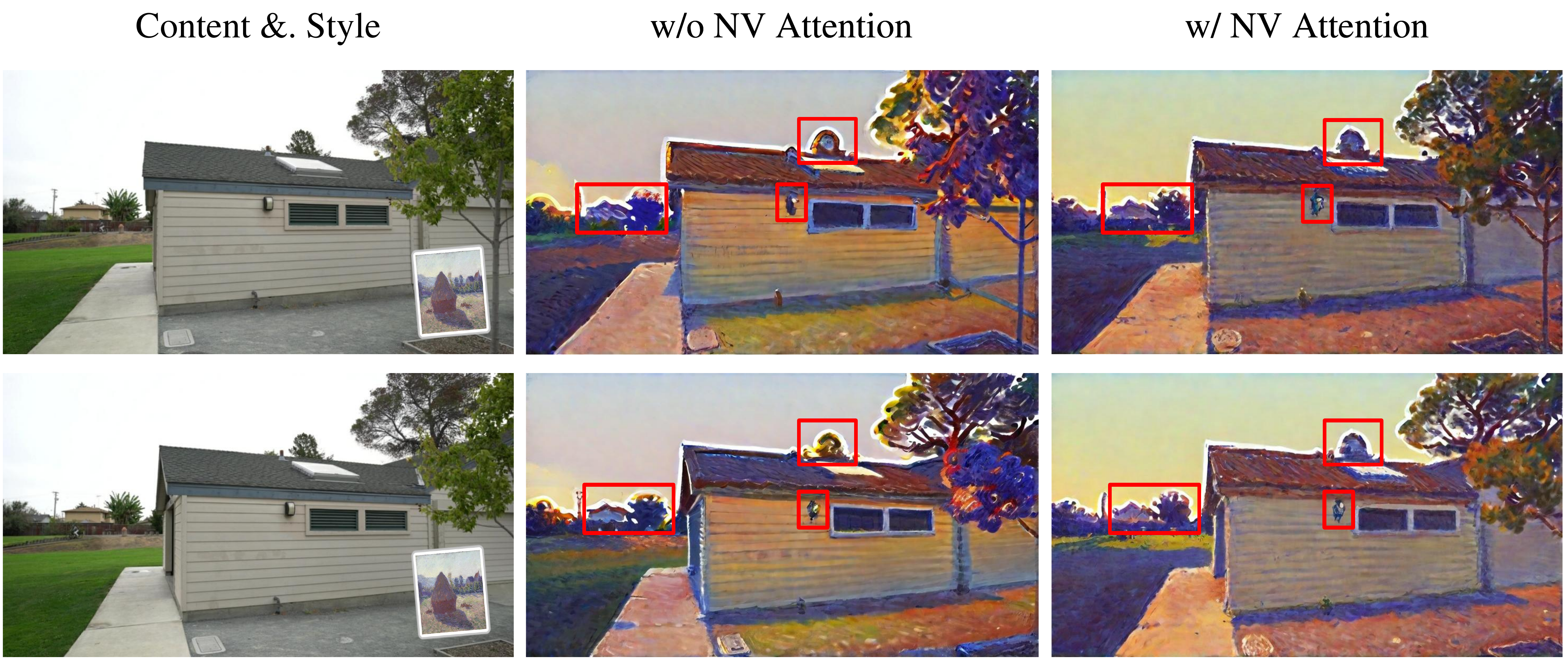}
	\caption{
		\textbf{Comparative Analysis of Detail Consistency in Neighboring Views with and without Neighboring-View Attention.}
		This illustration presents a comparative visualization of stylization results between two neighboring views. The regions demarcated by red bounding boxes clearly exhibit enhanced detail consistency when NV Attention is implemented, in contrast to the results obtained without NV Attention.}
	\label{fig:nv-contrast}
\end{figure*}

\
\newline
While our content control module offers effective content control, it is still difficult to ensure alignment among neighboring views.
Neighboring views refer to the views, of which camera centers are closer, meaning the corresponding images have larger overlapped regions.
Although neighboring views share identical style controls and similar content controls due to substantial region overlapping, their stylized outputs may demonstrate noticeable discrepancies among them.
This phenomenon triggers arbitrary fine-tuning in overlapping areas, resulting in color and texture discontinuities in detail across the scene.
Inspired by previous studies that achieve view consistency by manipulating attention \cite{liu2023text,wu2024gaussctrl,fujiwara2024style}, we propose an Attention-based Neighboring-View (NV) Alignment mechanism by modifying self-attention layers into Neighboring-View attention layers in the diffusion model.
Our approach demonstrates superior neighboring view consistency compared to the aforementioned approaches, as it employs attention sharing mechanism based on camera center distance rather than relying on random selection or predefined manual selection.
With attention sharing, images are no longer stylized alone.
Instead, images of neighboring views are jointly stylized as a group and as a result, details are more consistent among neighboring views (demonstrated in \cref{fig:nv-contrast}).

For $N$ neighboring views, let $Q_i$, $K_i$, $V_i$ be the queries, keys, and values of view $i$, and then we define the key and value of neighboring views as:
\begin{equation}\label{kv}
	K_{\mathrm{NV}}=[K_1, K_2, ..., K_N]^\mathsf{T}, V_{\mathrm{NV}}=[V_1, V_2, ..., V_N]^\mathsf{T}.
\end{equation}

Then, we denote the NV attention of view $i$ as:
\begin{equation}\label{attention}
	\mathrm{NVAttn}_i = \mathrm{Attn}(Q_i,K_\mathrm{NV},V_{\mathrm{NV}}),
\end{equation}
where $\mathrm{Attn}$ is the original self-attention.

We replace all self-attention layers with NV attention layers in the UNet module $\mathcal{F}_{\mathrm{UNet}}$ and the content control module $\mathcal F_c$ to get the Neighboring-View diffusion model.
We denote the NV features as $\hat z^t$.
Therefore, \cref{eq:prediction} becomes:
\begin{equation}
	\epsilon^t = \mathcal{F}_{\mathrm{UNet}}(z_c^t,\hat z^t,t,\mathcal F_c(z_c^t,\hat z^t,I_c, \mathcal D_s),\mathcal D_s).\label{eq:modified_prediction}
\end{equation}

\subsection{Implementation Details}
\label{Implementation-Details}
Our method is implemented by using the PyTorch library.
For 3D reconstruction and visualization, we use models and tools from the NeRFStudio \cite{tancik2023nerfstudio} library.
Specifically, we employ the ``splatfacto'' model for 3D reconstruction.
Following settings in CSGO \cite{xing2024csgo}, we employ Stable Diffusion XL v1.0 and pre-trained weight of CSGO models for 2D image stylization using the Diffusers \cite{von-platen-etal-2022-diffusers} library.

We conduct our experiments on an NVIDIA A800 GPU with 80GB of memory.
For the dataset update, we choose $N=15$ as the number of neighboring views.
It takes 15 minutes to complete the stylization of content images.
Our method includes a finetuning of 1,000 iterations, which costs 5 minutes.
In a word, our method only requires a total of 20 minutes to stylize the whole scene.
	\section{Experiment}
\label{sec:Experiment}
This section describes our experiments to evaluate our method.
We first introduce our experiment setup, including dataset, baselines, and metrics.
Then, we show both qualitative and quantitative results, followed by ablations of each proposed component.

\subsection{Experiment Setup}
\label{sec:exp-setup}
\textbf{Dataset.}
We evaluate our method on the Tanks and Temples dataset \cite{Knapitsch2017}, which includes unbounded $360^{\circ}$ real-world scenes with complex geometries and intricate details.
For style images, we use images from WikiArt \cite{wikiart}, allowing for evaluation of a wide range of artistic styles.

\noindent\textbf{Baselines.}
We compare our method with the state-of-the-art 3DGS stylization methods StyleGaussian \cite{liu2024stylegaussian} and InstantStyleGaussian \cite{yu2024instantstylegaussian}.
StyleGaussian is a feedforward-based framework, which embeds 3DGS and does stylization on the embeddings.
We use its pre-trained models for evaluation.
InstantStyleGaussian is also based on diffusion stylization, which can serve as an effective comparative baseline for evaluation.
We implement this method through a faithful reproduction of its original framework, adhering to its original specifications in its publication.

\noindent\textbf{Metrics.}
For content fidelity evaluation, we choose Content Feature Structural Distance (CFSD) \cite{chung2024style} as the metric, to mitigate the style influence in assessment.
To measure style similarity, we use Contrastive Style Descriptors (CSD) \cite{somepalli2024measuring} score.
Following previous diffusion-based 3D editing works \cite{fujiwara2024style,haque2023instruct}, we use CLIP Direction Consistency (CLIP-DC) to measure the temporal consistency across views.

\subsection{Qualitative Results}
\label{sec:qulitative-results}
\begin{figure*}[htbp]
	\centering
	\includegraphics[width=\linewidth]{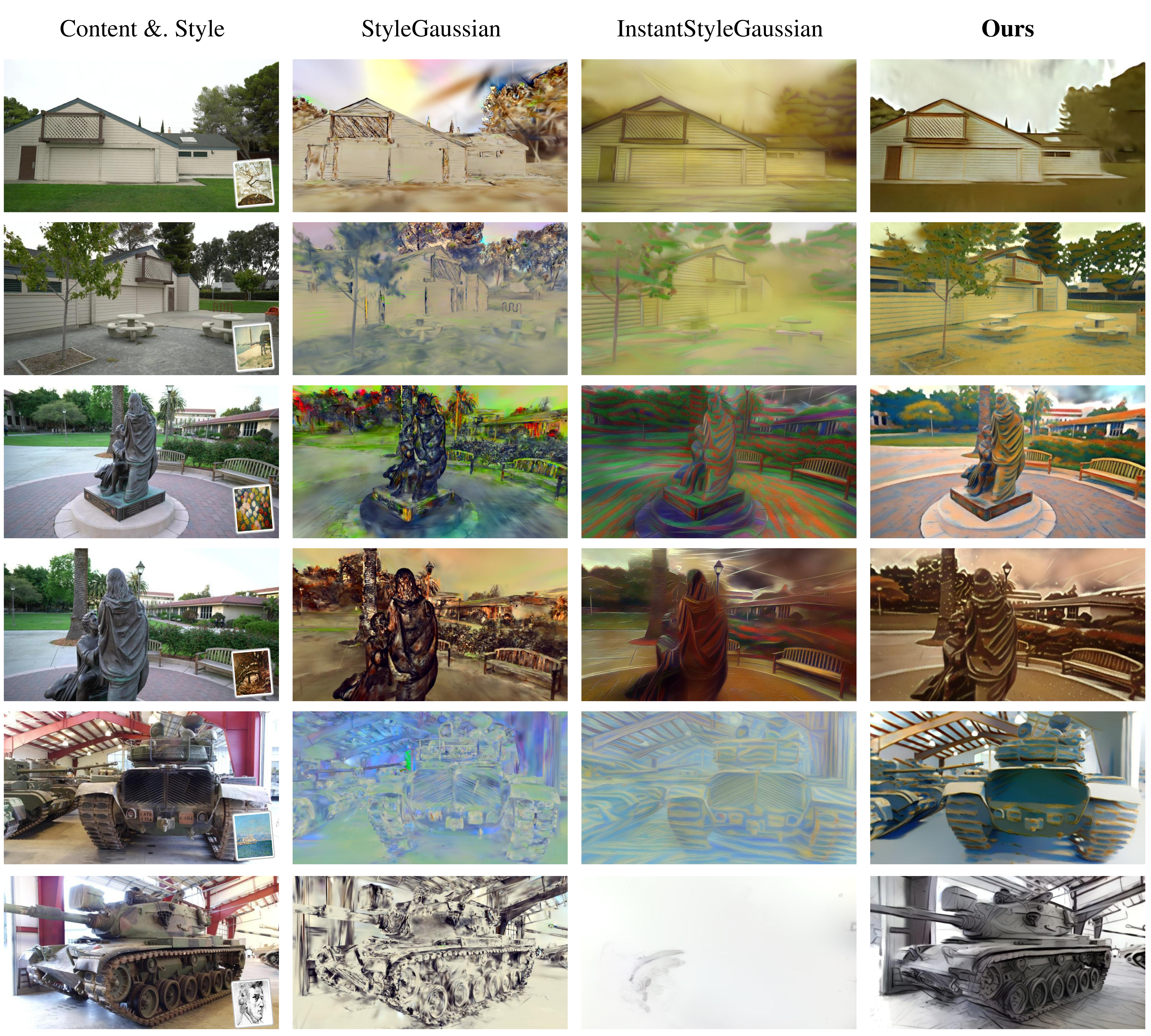}
	\caption{\textbf{Qualitative Results of Comparison.} We show diverse results of stylization in various scenes with different styles. Compared to SOTA methods StyleGaussian and InstantStyleGaussian, our method achieves the best visual quality by generating locally consistent scenes with better style alignment while preserving high content fidelity.}
	\label{fig:qr-fig}
\end{figure*}

We demonstrate the qualitative results in \cref{fig:qr-fig}, showcasing the advanced capabilities of ArtNVG in achieving superior style transfer performance.
ArtNVG excels in creating high-quality 3D scenes with enhanced alignment to the artistic style of reference images while preserving high content fidelity.
StyleGaussian struggles to prevent information leakage in rows 3 and 4, which is likely to originate from the style transfer process of embedded Gaussians.
In addition, StyleGaussian fails to generate 3D scenes with high quality, resulting in artifacts such as too many floaters (shown in row 1), blurring details (shown in rows 2 and 6), and locally inconsistent colors (shown in rows 3, 4, 5 and 6).
This phenomenon may be largely due to StyleGaussian's decoded stylized RGB design, by which the quality of the generated scenes depends on the decoded features.
Its features are extracted by VGG, resulting in the limitation of feature representation.
InstantStyleGaussian performs better than StyleGaussian in content preservation and visual quality.
However, it still suffers from limitations like blurring details (shown in rows 1, 2 and 4), severe content leakage (shown in row 4), and even occasional failure to reconstruct a meaningul 3D scene (shown in row 6), which may originates from its neglect of neighboring view consistency and adoption of an inversion-based content preservation approach in diffusion stylization.
It preserves the content information by adding noises on the original renderings following the implementation of some previous 3D editing frameworks \cite{haque2023instruct,dong2024vica,igs2gs,wu2024gaussctrl}, but this way of content preservation inherently poses a high risk of leaking the style information of content images.
Besides, it does not take the neighboring view consistency into consideration, causing locally arbitrary details in stylized images.
In contrast, our method uses Content-Style Separated Control to avoid information leakage and incorporates Attention-based Neighboring-View Alignment to promise locally consistent colors and textures, leading to fewer artifacts in scenes.

\subsection{User Study}
\label{sec:user study}
\begin{figure*}[htbp]
	\centering
	\includegraphics[width=0.8\linewidth]{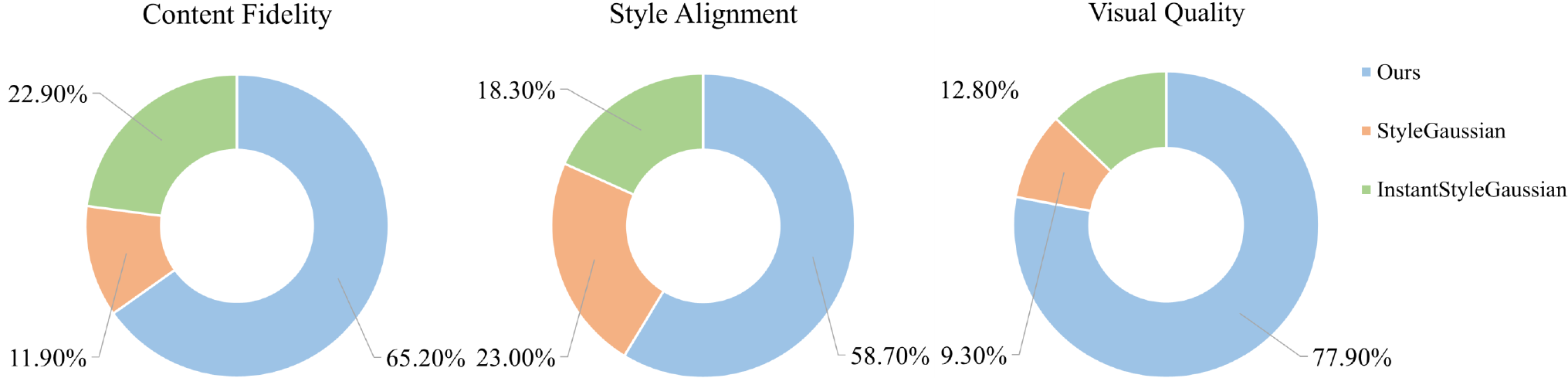}
	\caption{\textbf{User Study.} We record the user preference of our method and the baselines. Our method obtains more preference in content fidelity, style alignment, and visual quality than the baselines.}
	\label{fig:user study}
\end{figure*}
We conducted a user study comparing our method against StyleGaussian and InstantStyleGaussian.
We randomly chose 10 sets of stylized views rendered by different methods and invited 20 participants.
For each view, there are three evaluation indicators: content fidelity, style alignment and visual quality.
Then, participants were asked to vote for a view based on these indicators.
We collected a total of 200 votes for each indicator and demonstrated the results in \cref{fig:user study}.
Our method outperforms StyleGaussian and InstantStyleGaussian across all evaluation indicators, indicating our method can perform excellent artistic style transfer with high visual fidelity.

\subsection{Quantitative Results}
\label{sec:quantitative-results}
\begin{table}[t]
\centering
\resizebox{\columnwidth}{!}{

\begin{tabular}{lcccc}
\toprule
Method & CFSD ($\downarrow$) & CSD ($\uparrow$) & CLIP-DC ($\uparrow$)\\
\midrule
StyleGaussian & 0.26 & 0.11 & 0.77\\
InstantStyleGaussian & 0.23 & 0.10 & 0.79\\
Ours & \textbf{0.14} & \textbf{0.13} & \textbf{0.81}\\
\bottomrule
\end{tabular}
}
\caption{
\textbf{Quantitative Results of Comparison.} We evaluate the performance of our method against StyleGaussian and InstantStyleGaussian in terms of content fidelity, style similarity and temporal consistency, using CFSD ($\downarrow$), CSD score ($\uparrow$) and CLIP-DC ($\uparrow$). Our method surpasses them on all the metrics.
}
\vspace{-0.05in}
\label{tab:quantitative}
\end{table}

While the nature of stylization is inherently subjective, we mostly rely on the qualitative results and user study for evaluation.
Nevertheless, motivated by \cite{fujiwara2024style,haque2023instruct}, we apply auxiliary quantitative metrics over 8 scenes with different styles to evaluate our method more comprehensively.
The quantitative results are summarized in \cref{tab:quantitative}.
Our method outperforms StyleGaussian and InstantStyleGaussian on all evaluation metrics.
Particularly in content fidelity, our approach achieves significantly lower CFSD scores, compared to both StyleGaussian and InstantStyleGaussian, indicating enhanced capability in preserving structural content features. Regarding style alignment and multi-view consistency, our method obtains substantially higher CSD and CLIP-DC scores, surpassing the baselines.
These improvements primarily stem from the implementation of the Neighboring-View Attention mechanism, which effectively maintains cross-view consistency, and the integration of the CSGO model for enhanced feature representation.
In contrast, StyleGaussian relies solely on VGG features, while InstantStyleGaussian is limited to an inversion-based InstantStyle model, both of which lack effective design for maintaining view consistency, content fidelity and style alignment.

\subsection{Ablation Study}
\label{sec:ablations}
\begin{table}[t]
\centering
\resizebox{\columnwidth}{!}{
\begin{tabular}{lccc}
\toprule
Method & CFSD ($\downarrow$) & CSD ($\uparrow$) & CLIP-DC ($\uparrow$)\\
\midrule
Train from Scratch & 0.24 & 0.12 & 0.78\\
w/o NV Attention & 0.16 & 0.10 & 0.76\\
w/o NNFM Loss & \textbf{0.12} & 0.09 & 0.74\\
Ours & \textbf{0.12} & \textbf{0.15} & \textbf{0.80}\\
\bottomrule
\end{tabular}
}
\caption{
\textbf{Quantitative Results of Ablation Study.} The values are the average of the novel view renderings over four scenes using three styles. We use the same metrics in \cref{sec:quantitative-results} for evaluation.
}
\vspace{-0.05in}
\label{tab:ablations}
\end{table}

\begin{figure*}[htbp]
	\centering
	\includegraphics[width=\linewidth]{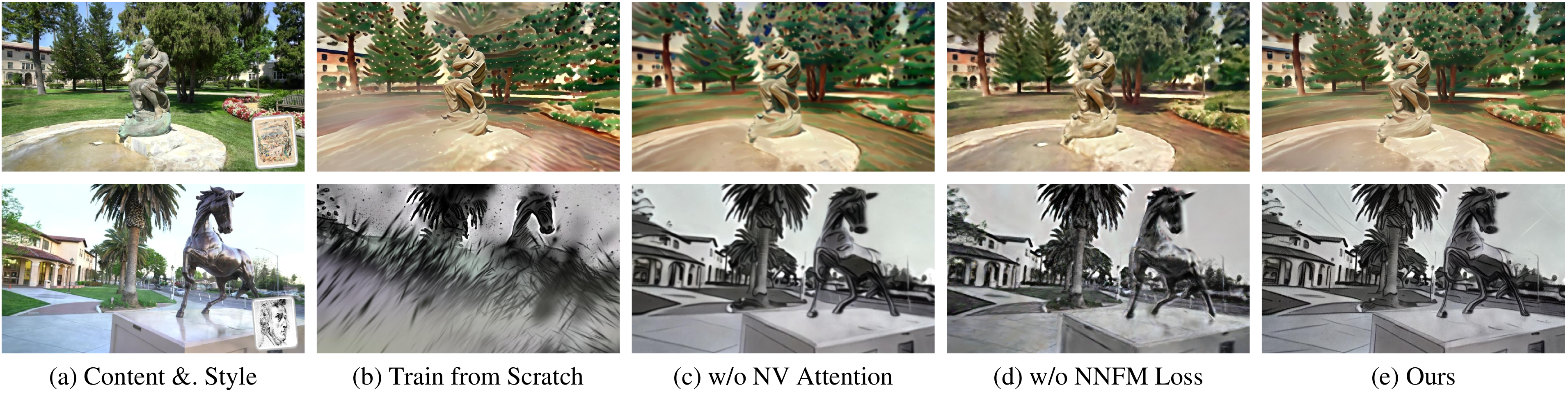}
	\caption{\textbf{Qualitative Results of Ablation Study.} We compare our method against several variants. The images show two example comparisons of the "Ignatius" and the "Horse" scenes with two different styles. Note that all images are novel view renderings from 3DGS. (b) is the result of training 3DGS scenes from scratch instead of optimizing the original ones. (c) displays the results by using the original diffusion model without Neighboring-View Attention. (d) demonstrates the result of optimization not using NNFM loss.}
	\label{fig:ablations}
\end{figure*}
\begin{figure}
	\centering
	\includegraphics[width=\linewidth]{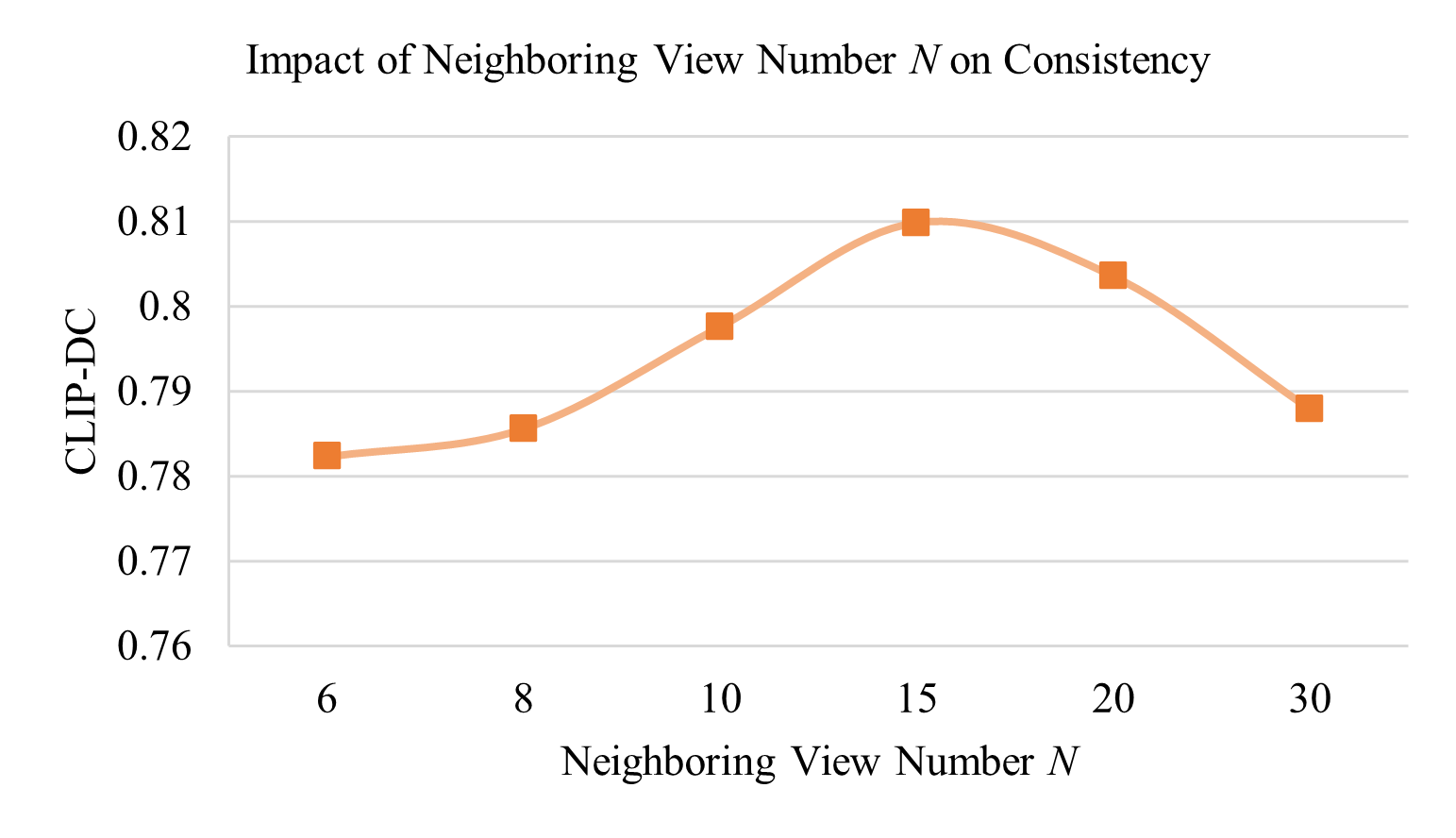}
	\caption{\textbf{Impact of the neighboring views number $N$ on consistency.} Comparing our method with different number of neighboring views $N\in\{6,8,10,15,20,30\}$, we evaluate the consistency across the whole 3D scene by CLIP-DC and find our method achieves the optimal consistency when $N=15$.}
	\label{fig:NV-num}
\end{figure}
We verify the effectiveness of our method by conducting an ablation study.
We compare our method to the following variants.
The qualitative differences are illustrated in \cref{fig:ablations}.
we also quantitatively measure our method against the variants using the metrics in \cref{sec:quantitative-results} and the results are demonstrated in \cref{tab:ablations}.
In addition, as shown in \cref{fig:NV-num}, we investigate the impact of the neighboring views number $N$ on consistency, which is an important hyperparameter in our method.

\noindent\textbf{Train from Scratch.}
In this simple variant, we train a 3DGS scene from scratch using stylized images generated by our Neighboring-View diffusion model.
As shown in \cref{fig:ablations} (b), the absence of a pre-trained geometric foundation of the 3D scene leads to several notable deficiencies in visual quality and fidelity, producing floating artifacts and inaccurate geometries due to ambiguities in geometry and color of stylized training images.
As a result, this variant also shows the highest CFSD, meaning its degree of content preservation is relatively the lowest.

\noindent\textbf{w/o NV Attention.}
To investigate the importance of maintaining the local consistency of stylized images, we conducted an ablation by disabling Neighboring-View attention.
As evidenced by in \cref{fig:ablations} (c), this variant harms the visual quality, leading to blurry details in finetuned 3DGS scenes, which is largely due to locally inconsistent details in neighboring views.
It can also be found that there is a decline in all metrics in \cref{tab:ablations}.

\noindent\textbf{w/o NNFM Loss.}
This variant finetunes the source 3DGS scenes just using the $L^1$ RGB loss without NNFM loss.
It can be observed in \cref{fig:ablations} (d) that finetuning with only RGB loss tends to cause artifacts such as splotchy Gaussians, blurry details, and locally inconsistent colors because stylized images generated with NV attention still have slight differences in detail.
Consequently, its CSD score and CLIP-DC are the lowest.
Thus, compared to RGB loss, NNFM loss is a more valid choice for eliminating artifacts and learning the style. 

\noindent\textbf{Neighboring View Number.}
The number of neighboring views significantly influences the consistency of the entire 3D scene.
We analyze the impact of varying neighboring views number, with $N \in \{6, 8, 10, 15, 20, 30\}$, on the overall scene consistency, which we measure using the CLIP-DC metric.
As shown in \cref{fig:NV-num}, both insufficient and excessive neighboring views lead to declines in consistency, with $N=15$ yielding the optimal consistency.
With insufficient neighboring views, the scene is prone to generate excessive clusters that fragment the whole 3D scene, disrupting the visual continuity among views.
Conversely, an excessive number of neighboring views causes the model to put too much emphasis on alignment in a single cluster, reducing the continuity across clusters.
Therefore, selecting an appropriate number of neighboring views is crucial as it ensures a balance between adequate clustering and maintaining continuity across clusters, thus enhancing the scene consistency.
	\section{Limitations}
\label{sec:limitations}

Our method, while achieving rapid and perceptually coherent 3D style transfer, presents certain limitations that may impact its application in more diverse contexts.
Our method primarily focuses on texture editing of scene surfaces and thus is unable to handle complex geometric transformations.
We believe that applying geometric deformation to stylization is a potential research field.
Additionally, our method does not integrate human interactions into the style transfer process, such as adding or removing objects from the scene through segmentation.
It may be possible to realize diverse interactions in stylization by incorporating various control modules of diffusion models, such as ControlNet.

	\section{Conclusion}
\label{sec:conclusion}
In this paper, we introduce ArtNVG, an innovative method for 3D stylization that excels in quickly generating a stylized 3D scene by specifying a reference style image.
ArtNVG significantly reduce the risk of information leakage in the style transfer process by using the Content-Style Separated Control.
By decoupling content and style control, our approach not only promises style alignment, but also ensures content fidelity. 
In addition, we propose an Attention-based Neighboring-View Alignment mechanism to solve the problem of local inconsistency, for our Neighboring-View diffusion model can generate stylized images with consistent textures and colors among neighboring views.
We have demonstrated that our method can produce high-quality results and outperforms the state-of-the-art method across a diverse set of scenes, showcasing its robustness and versatility.
The potential applications of ArtNVG extend beyond traditional media and entertainment domains, offering substantial benefits in game development, virtual reality, and augmented reality.
The ability to rapidly apply and modify styles in a consistent manner can greatly enhance user experiences and immersion.

	{\small
		\bibliographystyle{ieeenat_fullname}
		\bibliography{11_references}
	}
	
	\ifarxiv \clearpage \appendix \maketitlesupplementary This document serves as supplementary materials for \textit{ArtNVG: Content-Style Separated Artistic Neighboring-View Gaussian Stylization}, providing additional implementation details, metrics, and extended results. We highly recommend that reviewers view the \textbf{following results} in \cref{sec:app_add_res} to fully assess the performance of our stylization method across multiple perspectives.

\section{Additional Implementation Details}
\label{sec:app_implementation}

\begin{figure*}[htbp]
	\centering
	\includegraphics[width=0.85\linewidth]{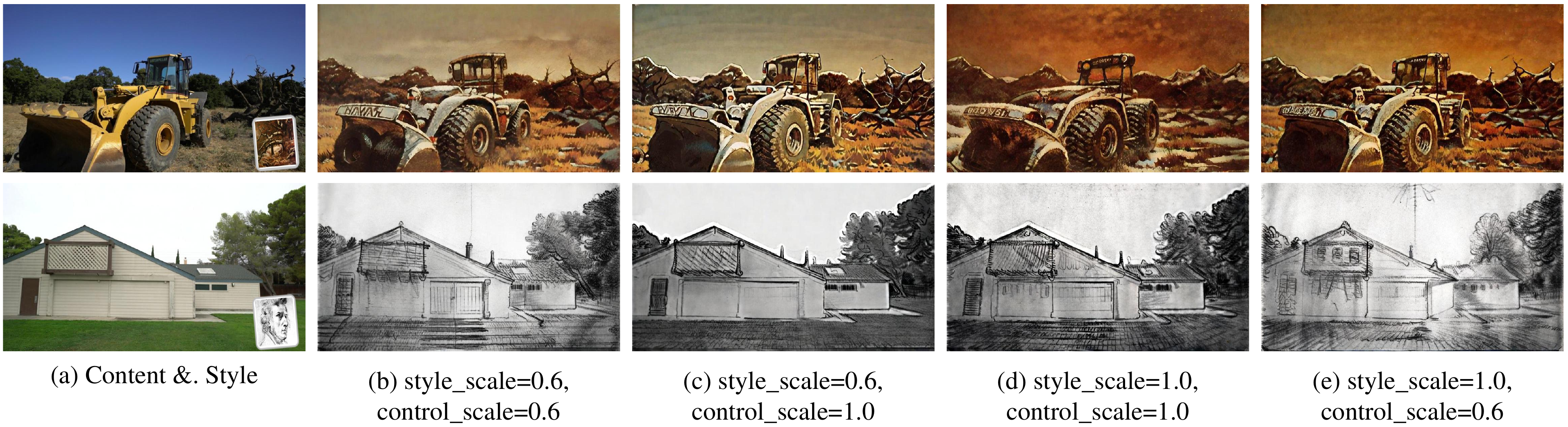}
	\caption{\textbf{Impact of $\mathbf{style\_scale}$ and $\mathbf{control\_scale}$ on content structure preservation.} We show stylized images using various combinations of hyperparameters  $(\mathrm{style\_scale},\mathrm{control\_scale})\in\{(0.6,0.6),(0.6,1.0),(1.0,1.0),(1.0,0.6)\}$. (c) achieves the best performance on content structure preservation.}
	\label{fig:app_1}
\end{figure*}

The primary input to our method is a 3DGS scene obtained by reconstruction.
To get this reconstruction, we pre-train the ``splatfacto" model from the NeRFStudio \cite{tancik2023nerfstudio} library for 20,000 iterations per scene.
Then, we finetune the 3DGS scene for 1,000 iterations to complete the stylization.
For per-training, we use the default ``splatfacto" losses, which mainly depend on the $L^1$ RGB loss, while we add the NNFM losses to enhance style alignment and details \cite{zhang2022arf} during finetuning. The finetuning loss function is denoted as:
\begin{equation}
	\mathcal L_\mathrm{fine} = \mathcal L_\mathrm{splatfacto} + \mathcal L_\mathrm{NNFM},
\end{equation}

Before finetuning, we update the whole dataset by stylizing the rendering images.
For dataset update, we have three hyperparameters and use
\begin{align}
	&\mathrm{content\_scale=1.0},\\
	&\mathrm{style\_scale=0.6},\\
	&\mathrm{control\_scale=1.0},
\end{align}
for all scenes.
Compared to the original setting of CSGO \cite{xing2024csgo} $(\mathrm{content\_scale},\mathrm{style\_scale},\mathrm{control\_scale})=(1.0,1.0,0.6)$, we inherit the value of $\mathrm{content\_scale}$, reduce the $\mathrm{style\_scale}$ and improve the $\mathrm{control\_scale}$ to preserve more structural information from the content image.
In \cref{fig:app_1}, we show two examples and images in a row are stylized with the same style image but different combinations of hyperparameters.
As we can see, both the rise of $\mathrm{style\_scale}$ and the reduction of $\mathrm{control\_scale}$ cause obvious declines in the content preservation and our setting achieves the best performance on preserving the structure from the content image.
The slight loss in the effect of stylization may be observed when we decrease the $\mathrm{style\_scale}$ and increase the $\mathrm{control\_scale}$.
However, it is more important to avoid significant geometry transformation, which severely harms the performance of finetuning due to a huge shift from the original geometry.

For the image encoder, we use IP-Adapter \cite{ye2023ip-adapter}.
For ControlNet, we use \textit{TTPLanet\_SDXL\_Controlnet\_Tile\_Realistic}.

\begin{figure*}[htbp]
	\centering
	\includegraphics[width=\linewidth]{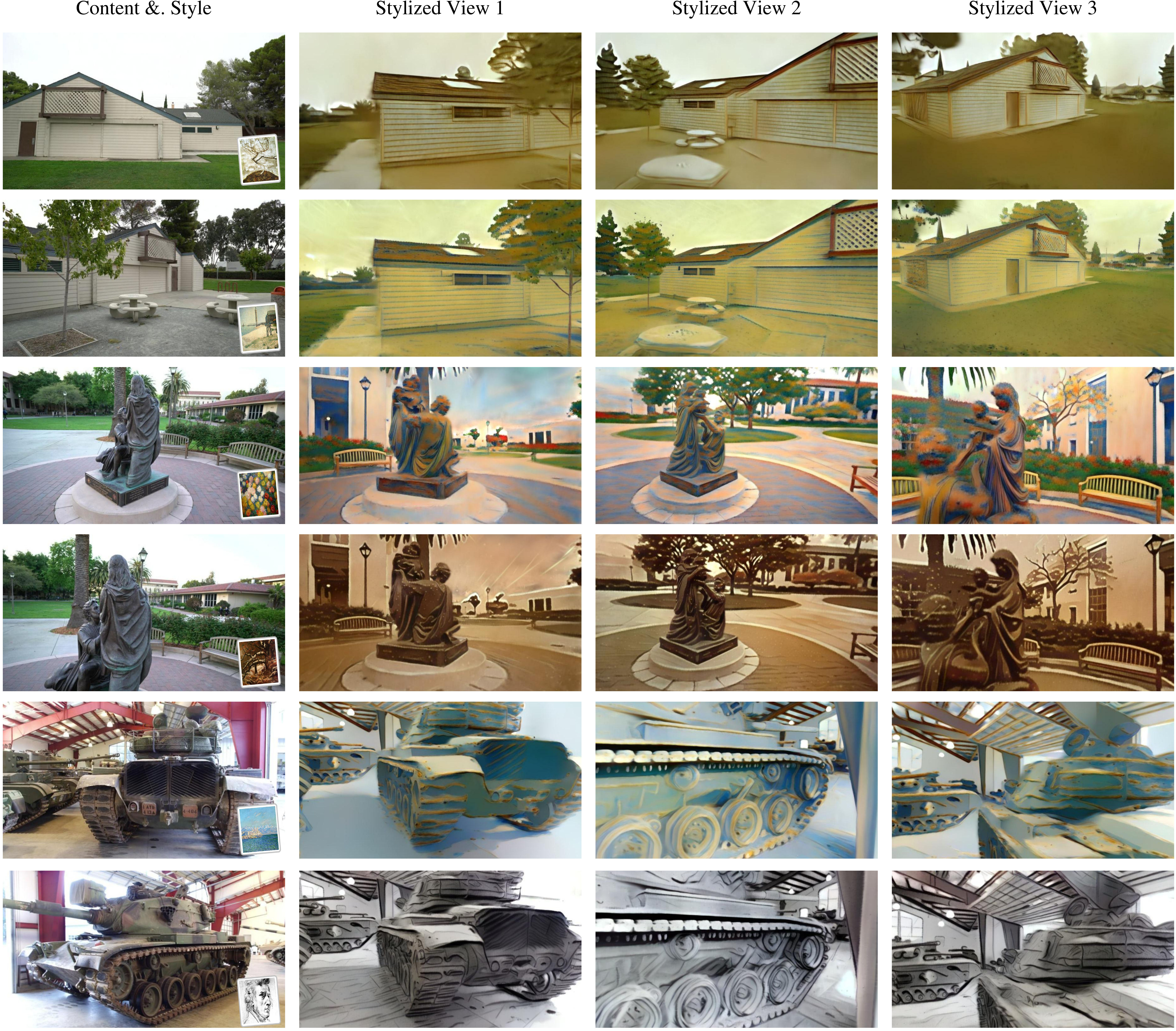}
	\caption{\textbf{Extended results.} We show three stylized views per scene.}
	\label{fig:app_2}
\end{figure*}

\begin{figure*}[htbp]
	\centering
	\includegraphics[width=\linewidth]{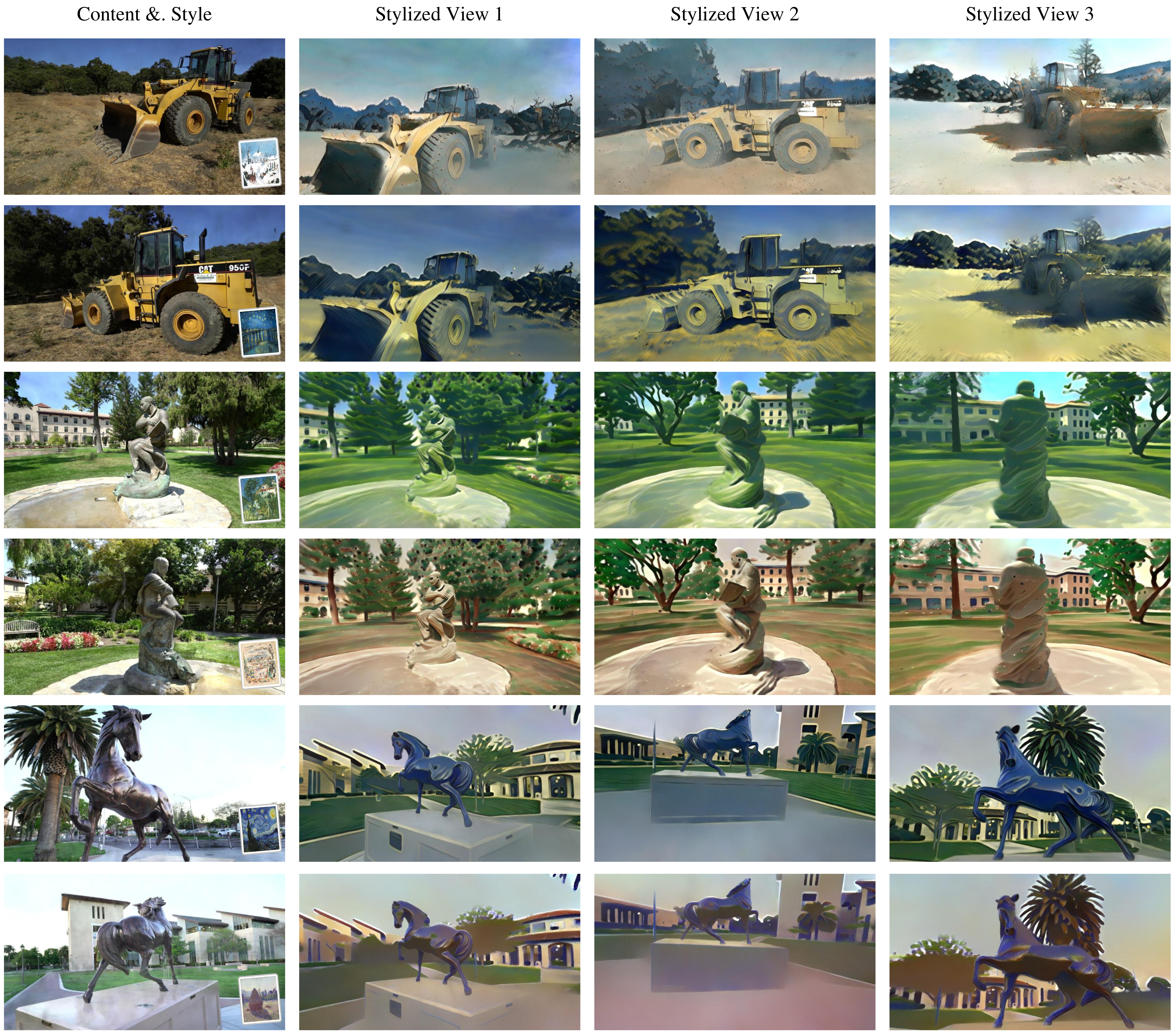}
	\caption{\textbf{Extended results.} We show three stylized views per scene.}
	\label{fig:app_3}
\end{figure*}

\section{Metrics}
\label{sec:metrics}
In this section, we give a detailed explanation of the metrics reported in the quantitative evaluation and the ablation study.

\noindent\textbf{Content Feature Structural Distance} 

Content Feature Structural Distance (CFSD) \cite{chung2024style} is a metric for evaluating content fidelity.
It only considers the spatial correlation between the original image and the stylized image, whereas the commonly used LPIPS distance may be affected by the style information.
For a given image $I$, we obtain its output feature of $conv3$ in VGG19 \cite{simonyan2014very} as feature maps $F\in \mathbb R^{hw\times c}$.
Then we define the patch similarity map as $M=F\times F^\mathsf{T}$ and the correlation map is represented as $S=[\mathrm{softmax}(M_i)]^{hw}_{i=1}$.
The CFSD between the correlation map of the content image ($S^c$) and the stylized image ($S^{cs}$) can be calculated as:
\begin{equation}
	\mathrm{CFSD}=\frac{1}{hw}\sum_{i=1}^{hw}\mathrm D_\mathrm{KL}(S^c_i||S^{cs}_i).
\end{equation}

\noindent\textbf{Contrastive Style Descriptors}

Contrastive Style Descriptors (CSD) \cite{somepalli2024measuring} is a model proposed to extract style descriptors from images, which has a good performance on the WikiArt \cite{wikiart} dataset used in our experiments.
Benefited from CSD's powerful capability of style extraction, we can calculate the style similarity of two images by the cosine similarity between their style descriptors output by CSD.
Given a style image $I_\mathrm{sty}$ and a stylized image $I_s$, the CSD score can be denoted as:
\begin{equation}
	\mathrm{D_{cos}}(\mathrm{CSD}(I_\mathrm{sty}),\mathrm{CSD}(I_s)),
\end{equation}
where $\mathrm{D_{cos}}$ is the cosine distance between two vectors and $\mathrm{CSD}(\cdot)$ is the output of CSD.

\noindent\textbf{CLIP Direction Consistency}

The CLIP Direction Consistency (CLIP-DC) score is proposed by \cite{haque2023instruct} to measure the cosine similarity of the CLIP embeddings of each pair of adjacent frames in a render novel camera path.
It can effectively measure the change of the editing direction in the CLIP-space from frame to frame.
Given $C(o_i)$, $C(o_{i+1})$, $C(s_i)$, $C(s_{i+1})$, four CLIP embeddings, corresponding to original 3DGS renderings $o_i$, $o_{i+1}$ and stylized 3DGS renderings $s_i$, $s_{i+1}$, the CLIP-DC score is defined as:
\begin{equation}
	\mathrm{D_{cos}}((C(s_i)-C(o_i)),(C(s_{i+1})-C(o_{i+1}))).
\end{equation}

\section{Extended Results}
\label{sec:app_add_res}
In order to evaluate our method more comprehensively, we provide multi-view images as extended results across different scenes with diverse styles in \cref{fig:app_2} and \cref{fig:app_3}. \fi
	
\end{document}